\documentclass[letterpaper, 10 pt, conference]{ieeeconf}  

\IEEEoverridecommandlockouts                              

\usepackage{amsmath} 
\usepackage{amssymb}  
\usepackage{cite}

\usepackage[utf8]{inputenc}    
\usepackage{bm}                
\usepackage{mathtools}         
\usepackage{graphicx}          
\usepackage{booktabs}
\usepackage{svg}
\usepackage{hyperref}

\title{\LARGE \bf
DYNAMO: Dependency-Aware Deep Learning Framework for Articulated Assembly Motion Prediction
}

\author{Mayank Patel$^{1}$, Rahul Jain$^{2}$, Asim Unmesh$^{2}$, and Karthik Ramani$^{1,2}$
\thanks{$^{1}$School of Mechanical Engineering, Purdue University, USA}%
\thanks{$^{2}$Department of Electrical and Computer Engineering, Purdue University, USA}%
}

\begin{document}
\maketitle
\thispagestyle{empty}
\pagestyle{empty}
\begin{abstract}
Understanding the motion of articulated mechanical assemblies from static geometry remains a core challenge in 3D perception and design automation. Prior work on everyday articulated objects such as doors and laptops typically assumes simplified kinematic structures or relies on joint annotations. However, in mechanical assemblies like gears, motion arises from geometric coupling, through meshing teeth or aligned axes, making it difficult for existing methods to reason about relational motion from geometry alone. To address this gap, we introduce \textbf{MechBench}, a benchmark dataset of 693 diverse synthetic gear assemblies with part-wise ground-truth motion trajectories. MechBench provides a structured setting to study coupled motion, where part dynamics are induced by contact and transmission rather than predefined joints. Building on this, we propose \textbf{DYNAMO}, a dependency-aware neural model that predicts per-part $\mathrm{SE}(3)$ motion trajectories directly from segmented CAD point clouds. Experiments show that DYNAMO outperforms strong baselines, achieving accurate and temporally consistent predictions across varied gear configurations. Together, MechBench and DYNAMO establish a novel systematic framework for data-driven learning of coupled mechanical motion in CAD assemblies.\href{https://dynamo-web.pages.dev/}{\textcolor{purple}{Project Page}}
\end{abstract}

\section{Introduction}



Articulated mechanical assemblies ranging from gear trains and linkages to cabinets are ubiquitous in engineered systems and everyday devices. For robots to interact with, manipulate, or even design such systems, it is crucial to understand how their constituent parts move. While modern CAD tools (e.g., SOLIDWORKS, Autodesk Fusion 360, Siemens NX) provide detailed 3D geometry, they rarely include executable motion specifications, limiting their utility for robotics applications \cite{gupta2024opening} such as assembly automation, motion planning \cite{mitra2010illustrating, chuyun2025application, yan2020rpm}, and simulation \cite{fu2024capt, liu2023category}. Bridging this gap requires methods that can infer physically valid part motions directly from static geometry.

To infer how parts move, a wide range of methods have been proposed that operate on inputs such as RGB-D images \cite{li2016mobility, abbatematteo2019learning, wang2024active}, 3D meshes\cite{sharf2014mobility, qiu2025articulate, hu2017learning}, or point clouds \cite{li2020category, liu2023semi, liu2023self}. These methods span supervised\cite{yi2018deep, abbatematteo2019learning}, self-supervised \cite{liu2023building, liu2023self}, and generative paradigms \cite{li2024dragapart, liu2024singapo, liu2024cage}, and typically aim to predict either the mobility of individual parts\cite{wang2019shape2motion} or the motion fields of articulated objects \cite{eisner2022flowbot3d, yan2020rpm, yi2018deep}. A wide range of methods have been proposed that rely on synthetic datasets with clean segmentation masks and joint information \cite{hu2017learning}, while other methods operate on real-world scans  \cite{liu2023paris, mao2022multiscan} to predict mobility, which is then used to derive motion. Broadly, these works estimates parts mobility parameters such as part axis, joint type, motion type, joint limits, learning deformation fields, or segmenting objects into movable components.

While prior works have explored various aspects of articulated object manipulation and motion understanding, most methods model kinematics by estimating the mobility of individual parts and then applying a fixed set of equations for motion generation \cite{liu2023semi, fu2024capt}. These approaches are effective when the motions of different parts are independent, where modelling individual mobilities often yields accurate predictions. However, in mechanical assemblies, part motions are inherently interdependent, and motion propagates from one component to another through complex coupling. Capturing this propagation is a fundamental yet underexplored aspect of articulated objects, and it is especially critical for accurately modelling and generating motions in mechanical assemblies.

To explore the problem of motion prediction under coupling in articulated assemblies, we focus on gear-based mechanisms such as rack-and-pinion systems and worm drives, which present unique challenges due to their interdependent part motions. In these systems, motion propagates through contact, meshing, and transmission, making it difficult for traditional models to reason about relational motion from geometry alone. Previous dataset have touched on coupling only in simple objects, as the construction of articulated CAD assemblies is labor intensive and requires detailed part modeling and articulation definitions~\cite{liu2025survey}. To address this gap, we introduce the MechBench, a novel dataset consisting of 693 synthetic gear assemblies with diverse configurations, providing a structured benchmark for predicting rigid-body motion in complex assemblies. Beyond the dataset, we propose \textbf{DYNAMO}, a dependency-aware neural model designed to address the coupling problem by predicting per-part SE(3) trajectories using 6D Lie algebra vectors. DYNAMO learns motion patterns directly from static, segmented CAD geometry, without relying on joint annotations or predefined kinematic structures, and integrates PointNet++ for part-level feature extraction, a graph neural network for modeling inter-part relationships, and a temporal decoder for trajectory prediction, enabling end-to-end learning of coupled rigid-body motion.

\noindent Our key contributions are:
\begin{itemize}
    \item We introduce MechBench, a novel benchmark dataset of 693 diverse synthetic gear assemblies with part-wise ground-truth motion labels, enabling systematic evaluation of coupled mechanical motion prediction.
    \item We present DYNAMO, a dependency-aware neural model that jointly reasons over inter-part relationships to predict per-part SE(3) motion trajectories from static CAD geometry.
    \item We compare against baselines and conduct ablation studies, demonstrating that DYNAMO achieves accurate and temporally consistent motion predictions on MechBench across multiple evaluations.
\end{itemize}

\section{Related Work}

Research on articulated object motion prediction has largely focused on everyday objects, aiming to infer part mobility or deformation fields from static geometry. Early works such as Shape2Motion~\cite{wang2019shape2motion} proposed joint discovery frameworks, while later methods explored supervised~\cite{yi2018deep, fu2024capt}, weakly-supervised~\cite{liu2023semi}, and self-supervised paradigms~\cite{liu2023self, shi2021self}. Other approaches model point-wise motion fields~\cite{yan2020rpm, eisner2022flowbot3d}, or track part poses given temporal cues~\cite{qian2022understanding}. These methods have driven progress but generally treat parts independently, assuming simple kinematic structures or relying on ground-truth joints.

To support these tasks, several synthetic datasets have been introduced. Early efforts such as Hu et al.~\cite{hu2017learning} and RPM-Net~\cite{yan2020rpm} provided motion annotations at limited scale, while Shape2Motion~\cite{wang2019shape2motion} and PartNet-Mobility~\cite{xiang2020sapien} expanded to thousands of objects across diverse categories. More recent datasets such as OPDSynth~\cite{sun2024opdmulti} and ACD~\cite{iliash2024s2o} target more realistic variations. However, these resources primarily capture independent part mobility and rarely include coupled motion. Table~\ref{tab:dataset_comparison} compares representative datasets, showing that MechBench is comparable in scale while uniquely addressing motion prediction under inter-part coupling in gear assemblies.

\begin{table}[h]
\centering
\caption{Comparison of synthetic articulated object datasets.}
\label{tab:dataset_comparison}
\resizebox{\linewidth}{!}{%
\begin{tabular}{lcccc}
\toprule
Dataset & \#Objects & \#Mov. Parts & \#Categories & Coupling \\
\midrule
Hu et al.~\cite{hu2017learning} & 368 & 368  & --  & -- \\
RPM-Net~\cite{yan2020rpm} & 969 & 1420 & 43  & Few \\
Shape2Motion~\cite{wang2019shape2motion} & 2440 & 6762 & 45  & -- \\
RBO~\cite{martin2019rbo} & 14  & 21   & 14  & -- \\
PartNet-Mobility~\cite{xiang2020sapien} & 2440 & 6762 & 45  & -- \\
OPDSynth~\cite{sun2024opdmulti} & 683 & 1343 & 11  & -- \\
ACD~\cite{iliash2024s2o} & 354 & 1350 & 21  & -- \\
\textbf{MechBench (Ours)} & \textbf{693} & \textbf{2445} & \textbf{13} & \textbf{Yes} \\
\bottomrule
\end{tabular}%
}
\end{table}

Mechanical assemblies present additional challenges beyond everyday articulated objects. Traditional kinematic modeling frameworks such as Mujoco~\cite{todorov2012mujoco} and Bullet~\cite{coumans2015bullet} assume known joints, while analysis methods~\cite{mitra2010illustrating, sharf2014mobility} attempt to infer mobility from contact. More recent learning approaches predict kinematic graphs from 3D data~\cite{abdul2022learning, liu2022toward}, but still rely on explicit joint representations. 
These assumptions break down for gears and transmissions, where motion propagates through meshing and contact rather than predefined joints. Addressing such coupled motion also depends on how rigid-body trajectories are represented: while quaternions or 6D continuous rotation formats~\cite{zhou2019continuity, weng2021captra} are widely used, they mainly target single-part pose estimation, and flow-based methods~\cite{yan2020rpm} lack rigid consistency. 
In contrast, our work adopts a Lie algebra formulation of SE(3), predicting per-part motion trajectories that are both valid and continuous, and enabling dependency-aware reasoning over coupled motions in complex assemblies.

\section{Dataset}

Gear-based assemblies such as rack-and-pinion mechanisms, planetary trains, and worm drives present an ideal setting for studying \emph{coupled motion} of parts, where the movement of one part directly induces motion in another through meshing and contact. Unlike everyday articulated objects (e.g., doors or laptops) with independent joints, gears capture the challenge of dependency-driven motion propagation. To enable learning in this setting, we construct \textbf{MechBench}, a benchmark dataset of synthetic CAD-generated gear assemblies with ground-truth motion labels.

\subsection{Assembly Creation}
\begin{figure}[ht]
\centering
\includegraphics[width=\linewidth]{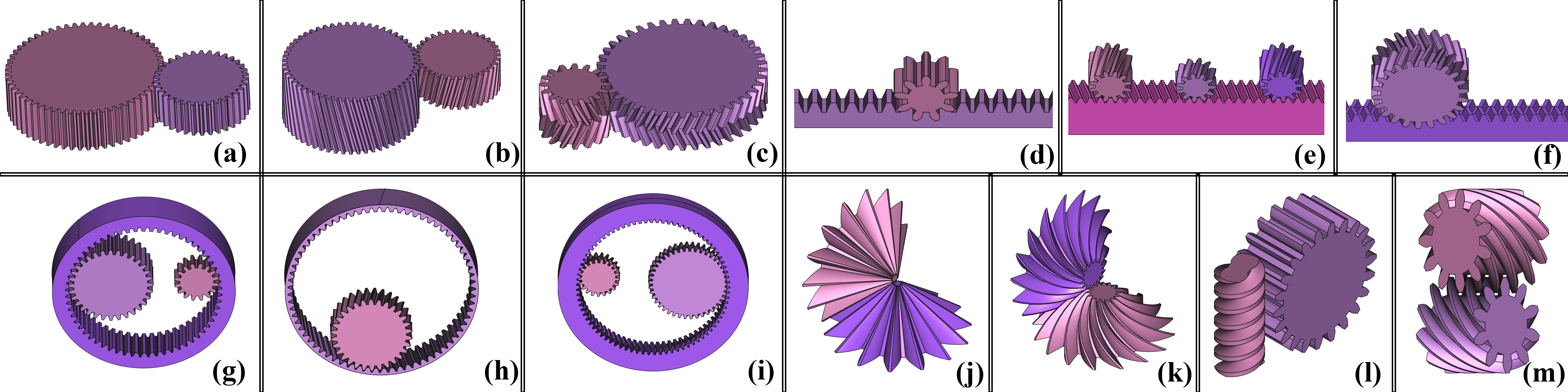}
\caption{Examples of diverse gear assemblies in our dataset, including: (a–c) spur and helical pairs, (d–f) planetary systems, (g–i) rack-and-pinion mechanisms, and (j–m) worm, bevel, and compound gear configurations.}
\label{fig:gear_examples}
\end{figure}

We procedurally generate assemblies using FreeCAD by varying key design parameters, including gear type (spur, helical, double-helical, bevel, worm, rack-and-pinion, planetary), number of teeth, diameter, pressure angle, and tooth profile. Figure~\ref{fig:gear_examples} illustrates examples across these categories. The number of parts per assembly is chosen randomly, ensuring valid meshing and mechanical compatibility. Each assembly is saved in two formats: (i) a \texttt{STEP} file preserving part hierarchy and geometry, and (ii) a segmented point cloud with up to $N_{\text{max}}=1024$ points per part, used for learning.

\subsection{Dataset Annotations}

Our dataset consists of comprehensive annotations for each assembly, including 3D meshes, segmentation masks, part-wise point clouds sampled uniformly from surfaces, frame-wise per part rigid-body motion annotations, and mobility units describing motion type, axis, and degrees of freedom.

\subsubsection{Motion Annotation}
For motion labels, we represent rigid-body motion using a 6D Lie algebra vector (6DLAV) $\boldsymbol{\xi} \in \mathbb{R}^6$, also known as a twist vector, derived from the Lie algebra $\mathfrak{se}(3)$ of the transformation group $\mathrm{SE}(3)$.  
\[
\boldsymbol{\xi} = [\boldsymbol{\omega}^\top, \mathbf{v}^\top]^\top, \quad 
\boldsymbol{\omega} \in \mathbb{R}^3, \; \mathbf{v} \in \mathbb{R}^3
\]
where $\boldsymbol{\omega}$ encodes angular velocity and $\mathbf{v}$ encodes linear velocity.

Rigid-body motions are traditionally represented as an $\mathrm{SE}(3)$ matrix, but direct regression of $R \in \mathrm{SO}(3)$ is unstable due to orthogonality and determinant constraints. Instead, 6DLAV provides a compact, continuous, and valid representation, easily converted to $\mathrm{SE}(3)$ via the exponential map:
\[
T = \exp\left(\hat{\boldsymbol{\xi}}\right) \in \mathrm{SE}(3)
\]

For each assembly, one gear is selected as the driver, rotating at $10^\circ$ per frame. The motion of other parts is determined by gear ratios. The ground-truth $\mathrm{SE}(3)$ transformation for part $k$ at frame $t$ is:
\[
T^{(k)}_{t} = T_{\text{center}}^{-1} \cdot R(\theta_t) \cdot T_{\text{center}}, \quad 
\theta_t = \theta_0 + t \cdot \Delta\theta^{(k)}
\]

\subsubsection{Mobility Annotation}
\begin{figure}[h]
\centering
\includegraphics[width=\linewidth]{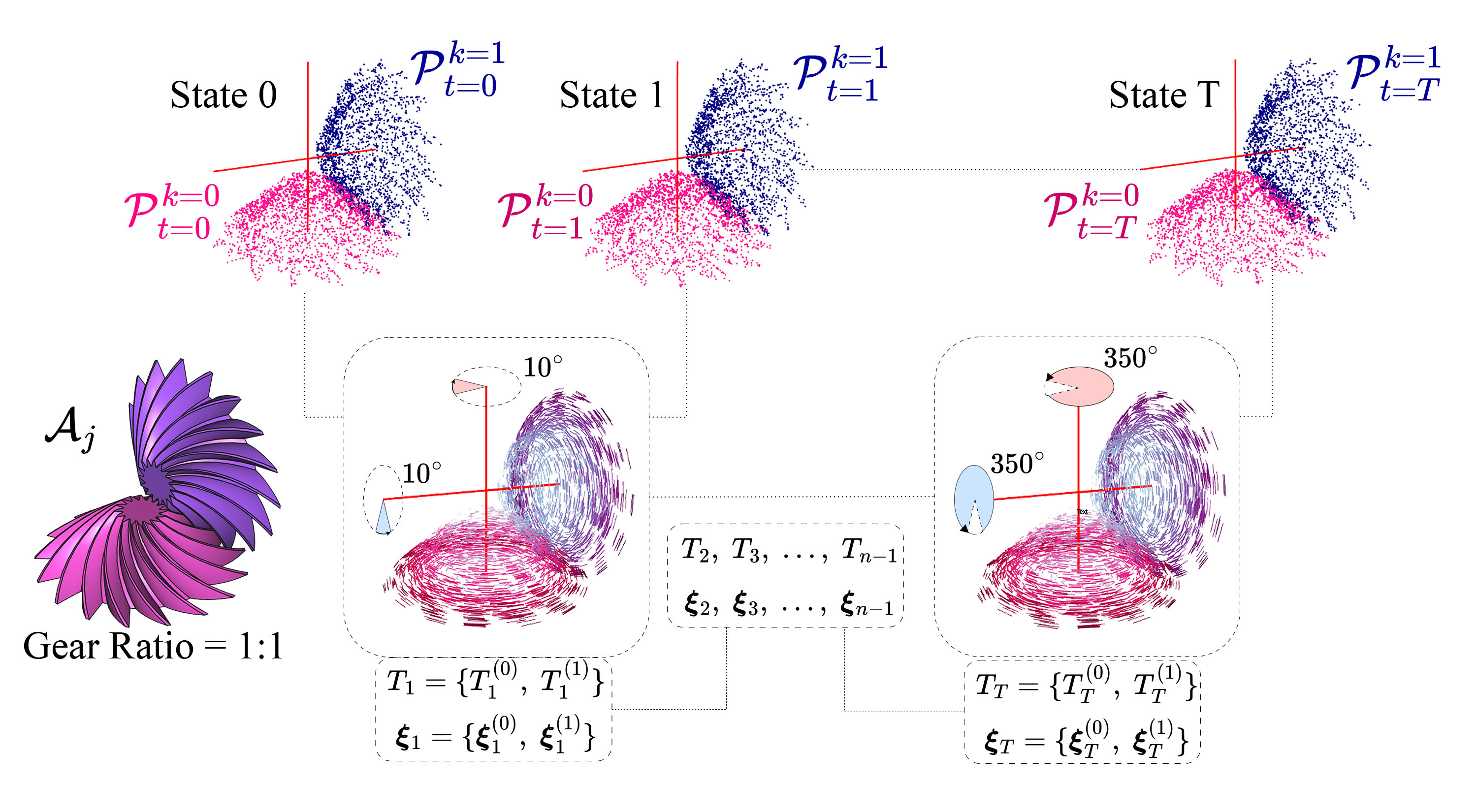} 
\caption{Illustration of a sample gear assembly $\mathcal{A}_j$ from our dataset, showing two-part meshing motion over time. For each part $k=0,1$, we visualize the segmented point cloud $\mathcal{P}^{(k)}_t$ at initial (State 0), intermediate (State 1), and final (State $T$) frames. Corresponding rigid-body motion is described by SE(3) transformations $T^{(k)}_t$ and twist vectors $\boldsymbol{\xi}^{(k)}_t$ at each time step $t$. A full motion sequence is composed of $T$ frames, where motion propagates according to gear ratio, and the dataset stores all $\{T_t\}_{t=1}^T$ and $\{\boldsymbol{\xi}_t\}_{t=1}^T$ for precise supervision.}
\label{fig:gear_datapoint}
\end{figure}

In addition, we store \emph{mobility units} for each part:
\[
\mu^{(k)} = (\tau^{(k)}, \mathbf{a}^{(k)}, \mathbf{c}^{(k)}, \mathbf{dof}^{(k)})
\]
where $\tau^{(k)} \in \{\text{Rotation}, \text{Translation}\}$ is the motion type, $\mathbf{a}^{(k)}$ is the axis, $\mathbf{c}^{(k)}$ the center, and $\mathbf{dof}^{(k)} \in \{0,1\}^6$ the active degrees of freedom. These units are used only for evaluation and not provided as input. Figure~\ref{fig:gear_datapoint} illustrates a sample two-part assembly, showing how point clouds evolve across frames together with their $\mathrm{SE}(3)$ transformations and twist vectors.

\subsection{Dataset Statistics}
MechBench contains 693 assemblies, denoted by $\mathcal{A}$. Each assembly $\mathcal{A}_j$ has $M_j$ parts, where $2 \leq M_j \leq 7$. The distribution is: 187 assemblies with 2 parts, 214 with 3 parts, 127 with 4 parts, 98 with 5 parts, 38 with 6 parts, and 29 with 7 parts. In total, the dataset provides 2445 movable parts. Each part is annotated across 36 frames of motion, corresponding to a complete rotation cycle that can be repeated for extended sequences.

\section{Method}

\begin{figure*}[t]
    \centering
    \includegraphics[width=\linewidth]{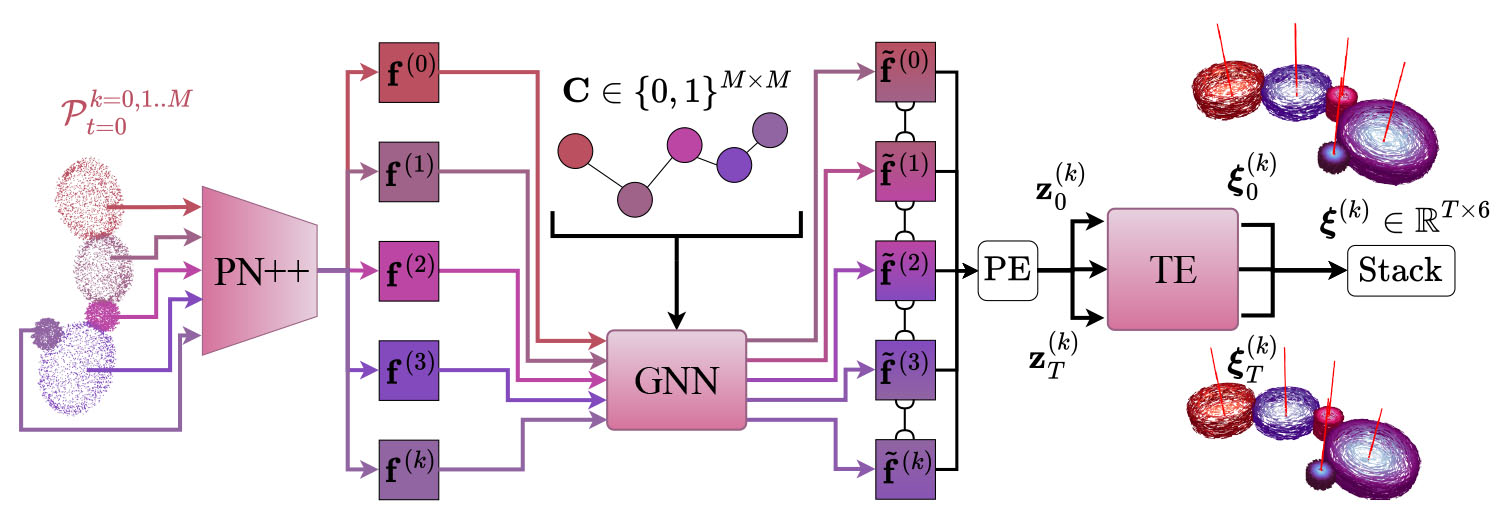}
    \caption{
    Overview of the proposed DYNAMO architecture. Given segmented part-wise point clouds $\mathcal{P}^{k=0,\dots,M}_{t=0}$, a PointNet++ backbone extracts per-part features $\mathbf{f}^{(k)}$, which are refined by a coupling-aware GNN based on a binary coupling matrix $\mathbf{C} \in \{0,1\}^{M \times M}$. The final features $\tilde{\mathbf{f}}^{(k)}$ are temporally replicated and added to positional encodings (PE) to form $\mathbf{z}_t^{(k)}$, which are fed into a Transformer decoder to predict twist vectors $\boldsymbol{\xi}_t^{(k)}$ over $T$ frames. The outputs are stacked to form the full motion trajectory $\boldsymbol{\xi}^{(k)} \in \mathbb{R}^{T \times 6}$.
    \textbf{Abbreviations:} PN++ = PointNet++, GNN = Graph Neural Network, PE = Positional Encoding, $\mathbf{C}$ = Coupling Matrix.
}
    \label{fig:DYNAMO_architecture}
\end{figure*}

\subsection{Problem Definition}

Given a CAD assembly of $M$ parts, the task is to predict the motion of all parts over $T$ frames. Input to the model is segmented part point clouds $\mathcal{P} = \{\mathcal{P}^k\}_{k=1}^{M}$, where each part $k$ is represented as $\mathcal{P}^k = \{\mathbf{p}_i^k\}_{i=1}^{N}$ with $\mathbf{p}_i^k \in \mathbb{R}^3$ as input and outputs a sequence of 6D Lie algebra vectors $\{\boldsymbol{\xi}^{(k)}_t\}_{t=1}^T$, where each $\boldsymbol{\xi}^{(k)}_t \in \mathbb{R}^6$ represents the motion of part $k$ at frame $t$ in twist form.

We assume that clean part-level segmentation is available since assemblies are generated directly from CAD models, making this a reasonable assumption. Unlike prior works such as Shape2Motion~\cite{wang2019shape2motion} and RPM-Net~\cite{yan2020rpm}, which must infer segmentation from raw scans, our setup focuses on learning motion reasoning rather than segmentation. While the CAD models also encode motion axes and centers, these quantities are withheld from the model and used only for evaluation.

To solve this task, we propose \textbf{DYNAMO} (Figure \ref{fig:DYNAMO_architecture}), a three-stage architecture consisting of (i) part-level point cloud feature extraction, (ii) coupling-aware inter-part relationship modeling, and (iii) temporal motion decoding. The following subsections describe each module in detail.

\subsection{Part Feature Extraction}

We extract geometric features for each part using a PointNet++ backbone with hierarchical Set Abstraction (SA) layers. The input tensor is $(P, 3, N)$, where $P$ the number of parts, and $N$ the number of points per part. Each part’s point cloud is independently encoded through three SA modules that apply farthest point sampling, neighborhood grouping, and shared MLP layers, followed by a global pooling stage. This hierarchical process captures local-to-global geometric context, and a final SA module produces a global feature vector $\mathbf{f}^{(k)} \in \mathbb{R}^C$ for each part, yielding a feature tensor of shape $(P, C)$ for downstream reasoning.




\subsection{Inter-Part Relationship Modeling}
In gear assemblies, motion of one part often directly induces motion in another due to meshing contact. These motion dependencies, referred to as \emph{coupling}, are fundamental to mechanical assemblies but are not always reflected in CAD file hierarchies. While some CAD tools store parent–child trees, these do not necessarily encode true motion propagation paths. Instead, coupling arises from geometric constraints, such as tooth interlocking or shaft alignment, which must be inferred from part geometry.

To capture such relationships, we model each assembly as a fully connected part graph with $M$ nodes, where each node corresponds to a part, and the edges encode mechanical coupling. Let $\mathbf{f}^{(k)} \in \mathbb{R}^C$ denote the feature vector of part $k$, obtained from the PointNet++ backbone, where $k = 1, \dots, M$. We define a binary \textbf{coupling matrix} $\mathbf{C} \in \{0,1\}^{M \times M}$, where $c_{ij} = 1$ indicates that parts $i$ and $j$ are coupled, and $c_{ij} = 0$ otherwise.

DYNAMO estimates the coupling matrix directly from geometry using a simple contact heuristic. For each pair of parts $(i, j)$, we compute all pairwise distances between their point clouds $\mathcal{P}^i$ and $\mathcal{P}^j$. If enough points lie closer than a distance threshold $\tau_d$, we treat the parts as mechanically coupled; otherwise, they are considered independent. Formally,
\[
c_{ij} =
\begin{cases}
1, & \text{if } \#\{(m,n): \|\mathbf{p}^i_m - \mathbf{p}^j_n\|_2 < \tau_d\} \;\geq\; \tau_c, \\
0, & \text{otherwise},
\end{cases}
\]
where $\tau_c$ is the minimum number of close contacts required. This process yields a symmetric binary coupling matrix $\mathbf{C} \in \{0,1\}^{M \times M}$ indicating which parts are in contact.

Given this matrix, we apply a coupling-aware graph neural network (GNN) where parts exchange information only along edges marked as coupled. At each layer, part $i$ aggregates messages from its coupled neighbors $j$ to update its representation:
\[
\tilde{\mathbf{f}}^{(i)} \leftarrow \mathbf{f}^{(i)} + \sum_{j=1}^{M} c_{ij} \cdot \mathrm{MLP}(\mathbf{f}^{(i)}, \mathbf{f}^{(j)}).
\]
Here $\tilde{\mathbf{f}}^{(i)}$ is the updated embedding of part $i$, enriched with information from its coupled neighbors. This process allows motion information to propagate through the contact graph.

\subsection{Temporal Motion Decoder}

The final part features $\tilde{\mathbf{f}}^{(k)} \in \mathbb{R}^C$ produced by the GNN are temporally replicated across $T$ frames and combined with learnable positional encodings to form per-frame tokens. These sequences are processed by a Transformer encoder, which models temporal dependencies across frames and produces an updated sequence of latent vectors. Each latent vector is then passed through a small MLP head to predict a 6D twist vector $\boldsymbol{\xi}_t^{(k)} \in \mathbb{R}^6$, representing the angular and linear velocity of part $k$ at frame $t$:
\[
\boldsymbol{\xi}_t^{(k)} = \mathrm{MLP}\!\left(\mathrm{Transformer}\!\left(\{\mathbf{z}_\tau^{(k)}\}_{\tau=1}^{T}\right)_t\right).
\]
where $\mathbf{z}_\tau^{(k)}$ denotes the positional-encoding–augmented token derived from $\tilde{\mathbf{f}}^{(k)}$. The final output  ($\mathbb{R}^{P \times T \times 6}$) representing per-part, per-frame twist predictions. This design enables the model to learn smooth, temporally coherent motion trajectories using global self-attention over time.






\subsection{Network Training and Loss Functions} \label{DYNAMOLosses} 

The DYNAMO model is trained to predict physically plausible motion trajectories over time, expressed as sequences of 6D twist vectors $\boldsymbol{\xi}_t^{(k)} \in \mathbb{R}^6$ for each part $k$ at frame $t$. During training, we supervise the model using ground-truth twist sequences derived from synthetic CAD assemblies, and optimize a multi-term loss function that balances trajectory accuracy and temporal consistency.

The training objective $\mathcal{L}_{\text{total}}$ consists of three loss components: L2 loss for translation, geodesic loss for rotation, and a consistency loss across frames. The total loss is defined as:

\begin{equation}
\mathcal{L}_{\text{total}} = \lambda_{\text{trans}} \mathcal{L}_{\text{trans}} + \lambda_{\text{rot}} \mathcal{L}_{\text{rot}} + \lambda_{\text{const}} \mathcal{L}_{\text{const}}
\end{equation}
where each $\lambda$ term is a scalar weight controlling the relative importance of each component.

\paragraph{Translation Loss}
We use an L2 loss to supervise the predicted linear velocity components $\mathbf{v}_t^{(k)} \in \mathbb{R}^3$ of each part:
\begin{equation}
\mathcal{L}_{\text{trans}} = \frac{1}{M} \sum_{k=1}^{M} \frac{1}{T} \sum_{t=1}^{T} \left\| \mathbf{v}_{t}^{(k)} - \mathbf{v}_{t}^{(k, \text{gt})} \right\|_2^2
\end{equation}
where $M_b$ is the number of valid (i.e., movable) parts in the $b$-th assembly and $T$ is the number of motion frames.

\paragraph{Rotation Loss}
To avoid ambiguities in angular velocity supervision, we adopt a geodesic rotation loss. Given angular velocity vectors $\boldsymbol{\omega}_{t}^{(k)}$, we form rotation matrices $R_{t}^{(k)} = \exp(\hat{\boldsymbol{\omega}}_{t}^{(k)})$ and $R_{t}^{(k,\text{gt})} = \exp(\hat{\boldsymbol{\omega}}_{t}^{(k,\text{gt})})$, and compute the loss as: 
\begin{equation}
\mathcal{L}_{\text{rot}} = \frac{1}{M} \sum_{k=1}^{M} \frac{1}{T} \sum_{t=1}^{T} \cos^{-1}\left( \frac{\mathrm{Tr}\left(R_{t}^{(k)\top} R_{t}^{(k, \text{gt})} \right) - 1}{2} \right)
\end{equation}

\paragraph{LAV Consistency Loss}
We encourage temporally smooth predictions by minimizing deviations in velocity and angular velocity over time. For each part $k$, we define frame-wise differences $\Delta \boldsymbol{\xi}_{t}^{(k)} = \boldsymbol{\xi}_{t+1}^{(k)} - \boldsymbol{\xi}_{t}^{(k)}$, and compute the variance-based loss
\[
\mathcal{L}_{\text{const}} = \frac{1}{M} \sum_{k=1}^{M} \mathrm{Var}_{t}\!\left(\Delta \boldsymbol{\xi}_{t}^{(k)}\right).
\]
This penalizes abrupt changes in part motion and encourages smoother, physically plausible trajectories.

\section{Evaluations}


\subsection{Experimental Setup}

Models are trained on MechBench, with a train/test split of 90/10. We use the AdamW optimizer with an initial learning rate of $10^{-4}$, weight decay of $10^{-5}$, and cosine annealing schedule with a minimum learning rate of $10^{-6}$. The model is trained with a batch size of 4, using gradient clipping at a norm of 1. All experiments are conducted on a single NVIDIA RTX A6000 GPU. All the models are trained with same hyperparameter for comparison. The loss function consists of three components: L2 translation loss, geodesic rotation loss, and a temporal consistency loss. We assign weights $\lambda_{\text{trans}} = 1$, $\lambda_{\text{rot}} = 1$, and $\lambda_{\text{const}} = 0.2$ to each component, which were empirically chosen via validation.

\subsection{Evaluation Metrics} \label{sec:metrics}

We evaluate predicted part motions using two metrics: \textit{rotation error} and \textit{translation error}. For both, we first transform each part’s point cloud $\mathcal{P}^k = \{\mathbf{p}_i^k\}_{i=1}^{N}$ using the predicted or ground-truth $\mathrm{SE}(3)$ trajectory $\{T_t^{(k)}\}_{t=1}^{T}$ to obtain $\mathbf{p}_{t,i}^k = T_t^{(k)} \cdot \mathbf{p}_i^k$. This yields frame-wise transformed point clouds $\mathbf{P}_t^{(k)}$ that allow direct comparison between predictions and ground truth.

\begin{table*}[h]
\centering
\caption{Rotation and translation errors on \textbf{MechBench}. Overall mean errors and breakdown by number of parts per assembly.}
\label{tab:overall_errors}
\setlength{\tabcolsep}{3pt} 
\resizebox{\linewidth}{!}{%
\begin{tabular}{lcccccccccccccc}
\toprule
& \multicolumn{2}{c}{Overall} & \multicolumn{2}{c}{2 Parts} & \multicolumn{2}{c}{3 Parts} & \multicolumn{2}{c}{4 Parts} & \multicolumn{2}{c}{5 Parts} & \multicolumn{2}{c}{6 Parts} & \multicolumn{2}{c}{7 Parts} \\
\cmidrule(lr){2-3} \cmidrule(lr){4-5} \cmidrule(lr){6-7} \cmidrule(lr){8-9} \cmidrule(lr){10-11} \cmidrule(lr){12-13} \cmidrule(lr){14-15}
Method & Rot & Trans & Rot & Trans & Rot & Trans & Rot & Trans & Rot & Trans & Rot & Trans & Rot & Trans \\
\midrule
RPM-Net~\cite{yan2020rpm} & 15.96 & 0.46 & 10.52 & 0.64 & 13.63 & 0.53 & 17.79 & 0.33 & 25.99 & 0.33 & 7.14 & \textbf{0.18} & 3.81 & \textbf{0.17} \\
RigidNet-LSTM & 9.64 & 0.64 & 6.04 & 0.69 & 9.09 & 0.66 & 9.29 & 0.58 & 16.08 & 0.65 & \textbf{5.04} & 0.64 & 3.76 & 0.41 \\
RigidNet-Transformer & 5.90 & 0.18 & 1.41 & 0.11 & 5.18 & 0.19 & 7.44 & 0.20 & 9.11 & 0.22 & 10.28 & 0.28 & 16.72 & 0.27 \\
\textbf{DYNAMO (Ours)} & \textbf{3.28} & \textbf{0.14} & \textbf{0.49} & \textbf{0.06} & \textbf{2.62} & \textbf{0.15} & \textbf{2.76} & \textbf{0.15} & \textbf{7.73} & \textbf{0.19} & 5.79 & 0.23 & \textbf{2.64} & 0.16 \\
\bottomrule
\end{tabular}%
}
\end{table*}

\paragraph{Rotation Error}
To quantify angular motion between frames, we compute the rigid rotation between the point clouds at consecutive frames using the Kabsch algorithm \cite{kabsch1976solution}. Specifically, let $\mathbf{P}_{t-1}^{(k)}$ and $\mathbf{P}_{t}^{(k)}$ be the transformed point clouds of part $k$ at frames $t-1$ and $t$, respectively. We center both point sets and apply the Kabsch algorithm to extract the relative rotation matrix $R_t^{(k)} \in \mathrm{SO}(3)$.

We compute the angular displacement in degrees as:
\[
\theta_t^{(k)} = \arccos\left( \frac{\text{trace}(R_t^{(k)}) - 1}{2} \right) \cdot \left( \frac{180}{\pi} \right)
\]
The \textit{rotation error} is the absolute difference in angular displacement between predicted and ground-truth sequences:
\[
\text{RotationError}_t^{(k)} = \left| \theta_t^{(k,\text{pred})} - \theta_t^{(k,\text{gt})} \right|
\]

\paragraph{Translation Error}
Translation error is computed as the mean Euclidean distance between the predicted and ground-truth transformed point clouds at each frame:
\[
\text{TranslationError}_t^{(k)} = \frac{1}{N} \sum_{i=1}^{N} \left\| \mathbf{p}_{t,i}^{k,\text{pred}} - \mathbf{p}_{t,i}^{k,\text{gt}} \right\|_2
\]

We evaluate both metrics per part $k = 1, \dots, M$ and per frame $t = 1, \dots, T$. We then report the average rotation and translation error over all movable parts and all frames across the test set. This evaluation framework is compatible with models that predict absolute $\mathrm{SE}(3)$ transforms (e.g., via 6D Lie Algebra vectors $\boldsymbol{\xi}_t^{(k)}$) and those that predict frame-wise point displacements (e.g., RPM-Net).

\subsection{Baselines}\label{baselines}
There are no existing methods designed for dependency-aware motion prediction in CAD assemblies. To provide fair comparisons, we introduce two simple yet strong baselines: \textbf{RigidNet-LSTM} and \textbf{RigidNet-Transformer}. Both take segmented part point clouds as input, extract features with PointNet++ (PN++), and use either an LSTM or Transformer temporal head to predict per-part 6D Lie algebra vectors over all frames. RigidNet-LSTM is inspired by RPM-Net’s architecture, but instead of predicting dense displacement flows, it directly outputs rigid-body motions per part, making it a natural recurrent baseline. In contrast, RigidNet-Transformer leverages self-attention to capture long-range temporal dependencies, offering a stronger alternative to recurrent models. Unlike DYNAMO, these models have no coupling supervision, allowing us to isolate the effect of temporal modelling alone. They are trained with the same loss functions as DYNAMO (Section~\ref{DYNAMOLosses}) for direct comparison.  


We also compare with \textbf{RPM-Net}~\cite{yan2020rpm}, which remains the only prior method capable of predicting multi-part motion trajectories from 3D geometry, for multiple frames of motion. Although RPM-Net jointly infers segmentation and motion, we disable its segmentation branch and use ground-truth part labels to focus on motion quality. Other existing methods target different settings (e.g., single-part mobility prediction, joint inference), making RPM-Net the most relevant baseline for our task.

\subsection{Quantitative Results}

We report rotation and translation errors for all baselines and DYNAMO. Table~\ref{tab:overall_errors} summarizes the overall mean errors across the entire test set. DYNAMO achieves the lowest errors in both metrics, reducing rotation error by more than $70\%$ compared to RPM-Net and consistently outperforming the simple rigid baselines. 

We also break down performance by number of parts in the assembly (Table~\ref{tab:overall_errors}). While all methods show increased difficulty for assemblies with more parts, DYNAMO maintains the lowest error across all settings, especially for challenging 5–6 part assemblies.

\paragraph{Frame-wise stability.}  
For rotation, errors remained stable across the full 35 frames with negligible fluctuations. The standard deviation (SD) of frame-wise rotation errors was $0.007^\circ$ for DYNAMO, $0.006^\circ$ for RPM-Net, $0.005^\circ$ for RigidNet-LSTM, and $0.062^\circ$ for RigidNet-Transformer, confirming that the averages reported in Table~\ref{tab:overall_errors} are consistent throughout the sequence.  

For translation, small temporal variations are observed, DYNAMO achieved the lowest SD ($0.058$), followed by RigidNet-Transformer ($0.051$), RPM-Net ($0.099$), and RigidNet-LSTM ($0.175$). These trends reinforce that DYNAMO not only yields the lowest overall translation errors but also maintains stable performance across time, while the other baselines exhibit larger fluctuations that compound their higher mean errors.


\begin{figure}[t]
  \centering
  \includegraphics[width=\linewidth]{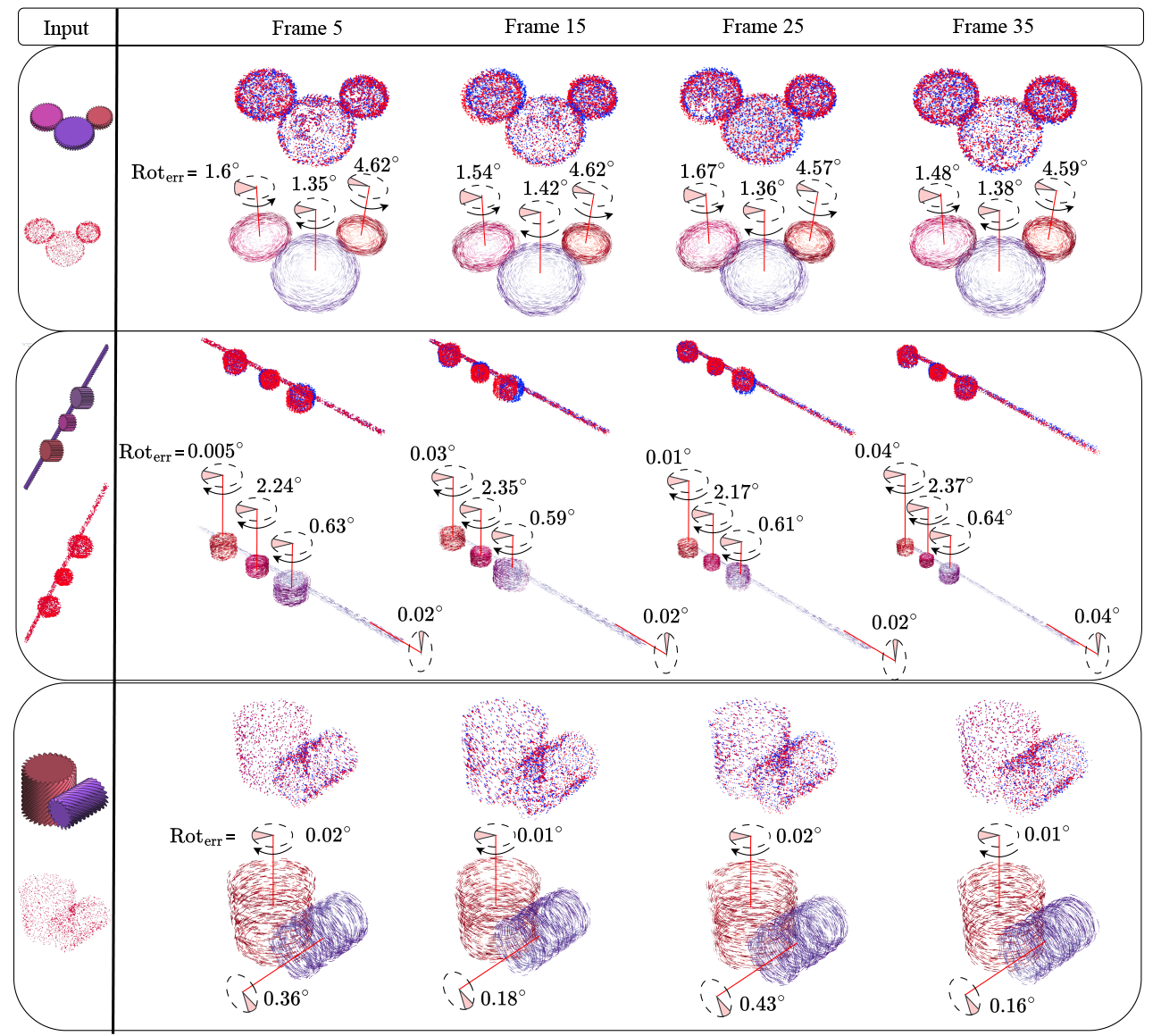}
  \caption{
  Qualitative motion prediction results across different CAD assemblies. Each pair of rows shows a test assembly with its input (top-left: reference mesh + segmented input point cloud), followed by predicted motion at Frames 5, 15, 25, and 35. In each frame, \textcolor{blue}{\textbf{blue}} denotes ground-truth point cloud, and \textcolor{red}{\textbf{red}} denotes the predicted point cloud overlaid. The bottom row for each frame shows the \textbf{point-wise displacement vectors} from the previous frame (e.g., Frame $t-1 \rightarrow t$), visualizing local motion trends. Annotations indicate the rotation direction of parts and report the exact angular error at each frame, highlighting prediction accuracy over time. Assemblies include spur, rack-and-pinion, and screw gears, demonstrating DYNAMO’s generalization to diverse, tightly coupled kinematic structures.
  }

  \label{fig:Results}
\end{figure}

\subsection{Qualitative Results} \label{sec:qualitative}


Figure~\ref{fig:Results} shows that \textbf{DYNAMO} accurately predicts both rotational and translational motions across diverse assemblies. It captures synchronized rotation in coupled planar gears, infers linear translation in rack-and-pinion systems, and transfers motion across orthogonal or misaligned axes in screw and bevel configurations. In all cases, DYNAMO maintains tight alignment with the ground truth and produces smooth, consistent displacement trajectories, demonstrating robustness even in geometrically complex or symmetric setups where simpler models often fail.

The predictions preserve contact constraints and rotational directionality (clockwise vs. counterclockwise), even when the axes of rotation are orthogonal or misaligned, as in bevel and screw gears. Despite no supervision on gear ratios or explicit axis labels, DYNAMO learns to internalize speed and coupling constraints from geometry alone. This includes understanding how the size of interacting gears affects their relative motion: when a small gear meshes with a larger one, the smaller gear must rotate faster to maintain contact, this mechanical relationship is known as the gear ratio. For example, if one gear is twice the diameter of its neighbor, the smaller gear will rotate twice as fast. DYNAMO appears to infer such relationships directly from the part geometries, learning that larger differences in gear sizes imply faster relative speeds, whereas similar-sized gears (e.g., ~1:1 ratio) rotate at similar speed. This emergent behavior indicates the model’s capacity to implicitly capture physical constraints purely from static 3D geometry.

Overall, these results show that DYNAMO not only predicts physically plausible part-wise motions but also captures coupling-induced behaviors in complex, multi-part CAD assemblies across 2 to 7 parts.

\subsection{Failure Cases} \label{sec:failures}
Although \textbf{DYNAMO} performs robustly across diverse assemblies, it can fail when coupling constraints are not preserved. This typically occurs because the model learns motion patterns only from geometry, without explicit physical priors, and may therefore decouple translation from rotation or allow parts to drift, leading to unphysical behaviors such as overlaps or disconnected motion. Such errors remain uncommon: across the 70 held-out assemblies, only one exceeded a $40^\circ$ rotation error and just two crossed $10^\circ$, with the vast majority below $5^\circ$.

\begin{figure}[h]
  \centering
  \includegraphics[width=0.8\linewidth]{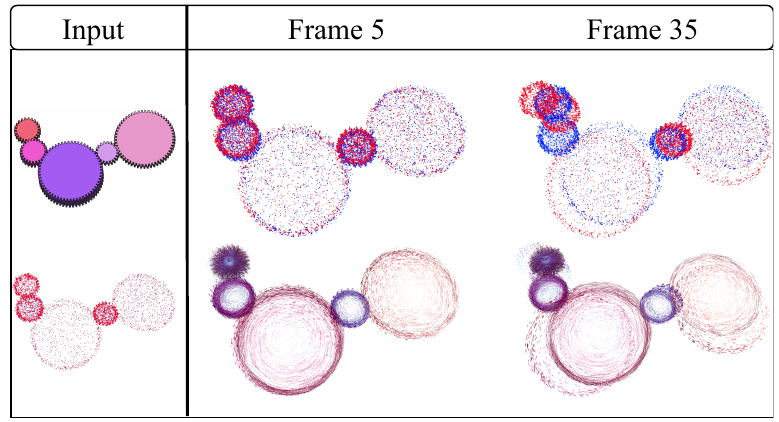}
  \caption{Edge case example with the highest observed rotation error ($49.67^\circ$). Red = predicted point cloud; Blue = ground truth. Bottom row shows displacement vectors (from previous frame) to visualize per-point motion direction.}
  \label{fig:EdgeCase}
\end{figure}

Figure~\ref{fig:EdgeCase} illustrates the most severe failure. In this five-gear assembly, the prediction diverges from the ground truth, producing merged and drifting parts and a gear that translates without rotation. This example highlights the model’s limitations in preserving strict coupling, suggesting future improvements such as incorporating contact-aware priors or explicit constraint enforcement during inference.

\subsection{Ablation Studies} \label{sec:ablation}

We conduct a series of ablation studies to assess the impact of key architectural and training choices in DYNAMO. Our goal is to quantify the contribution of different modules and design decisions toward accurate and stable motion prediction.

\paragraph{Model Architecture: Point Feature Extractor and Temporal Module.}
We vary the point cloud encoder (PointNet++ \cite{qi2017pointnet++} vs. Point Transformer \cite{zhao2021point}) and temporal head (LSTM vs. Transformer) while keeping the coupling-aware GNN fixed (Table~\ref{tab:ablation_architecture}). DYNAMO (PN++ + Transformer) achieves the best rotation accuracy ($3.32^\circ$) while remaining competitive in translation. Point Transformer with a Transformer head yields the lowest translation error ($0.136$), reflecting its strength in local geometric detail, but performs worse in rotation, where PN++’s hierarchical abstraction better captures global shape context. LSTM-based variants consistently degrade performance, showing that recurrent models struggle with long-horizon consistency compared to temporal self-attention. Overall, these results highlight that PN++ provides robust global features for rotation, while the Transformer decoder ensures stable trajectories across time, explaining the advantage of DYNAMO’s chosen design.

\begin{table}[h]
\centering
\caption{Ablation on point feature extractor and temporal module.}
\label{tab:ablation_architecture}
\begin{tabular}{lcc}
\toprule
Architecture Variant & Rotation ($^\circ$) & Translation \\
\midrule
Point Transformer + LSTM & 4.3014 & 0.1597 \\
Point Transformer + Transformer & 3.4063 & \textbf{0.1357} \\
PointNet++ + LSTM & 3.3761 & 0.1897 \\
\textbf{PointNet++ + Transformer}& \textbf{3.2822} & 0.1427 \\
\bottomrule
\end{tabular}
\end{table}

\paragraph{Effect of Point Cloud Density.}
We evaluate DYNAMO under varying levels of point cloud sparsity by randomly subsampling each part to 128, 256, 512, or 1024 points. As shown in Table~\ref{tab:ablation_density}, increasing point density improves both rotation and translation accuracy, with performance stabilizing around 512–1024 points. Notably, even at low resolutions (e.g., 256 points), DYNAMO achieves lower errors than all baselines tested in Section~\ref{baselines} (Table~\ref{tab:overall_errors}), demonstrating strong robustness to incomplete or partial point clouds. This makes our model suitable for real-world use cases where noisy or partial observations are common, such as depth sensing, occlusion-prone environments, or imperfect CAD imports.

\begin{table}[h]
\centering
\caption{Ablation on number of points per part point cloud.}
\label{tab:ablation_density}
\begin{tabular}{lcc}
\toprule
Points per Part & Rotation ($^\circ$) & Translation \\
\midrule
128 & 7.1212 & 0.4371 \\
256 & 3.7709 & 0.2076 \\
\textbf{512} & \textbf{3.1742} & 0.1474 \\
1024 & 3.2822 & \textbf{0.1427} \\
\bottomrule
\end{tabular}
\end{table}

\paragraph{Effect of Loss Terms.}
We ablate each loss component from the training objective to assess its individual contribution. Removing the rotation loss leads to a dramatic degradation in both metrics, especially rotation ($12.89^\circ$), confirming its critical role in orientational accuracy. Without translation loss, the model still learns rotational trends but suffers a large increase in translation error. Lastly, removing the temporal consistency loss results in moderate error increases, and more importantly, introduces temporal instability. The standard deviation (SD) of frame-wise translation error rises from $0.058$ (full model) to $0.080$, indicating less smooth trajectories over time.

\begin{table}[h]
\centering
\caption{Ablation on loss components. Removing any term degrades performance.}
\label{tab:ablation_losses}
\begin{tabular}{lcc}
\toprule
Loss Configuration & Rotation ($^\circ$) & Translation \\
\midrule
w/o Translation Loss & 3.7301 & 0.6043 \\
w/o Rotation Loss & 12.8936 & 2.5456 \\
w/o Temporal Consistency Loss & 3.4711 & 0.1586 \\
\textbf{All Losses (Ours)} & \textbf{3.2822} & \textbf{0.1427} \\
\bottomrule
\end{tabular}
\end{table}

\section{Conclusion and Future Work}

We presented \textbf{DYNAMO}, a dependency-aware neural model for predicting per-part rigid-body motion in CAD assemblies, together with \textbf{MechBench}, a benchmark of 693 synthetic gear assemblies annotated with ground-truth trajectories. Our results show that DYNAMO learns to infer coupled motions directly from static geometry, consistently outperforming strong baselines across multiple metrics.  

Despite these advances, the model relies solely on geometric cues and can produce implausible behaviors such as drifting or intersecting parts. Future work will investigate incorporating physical constraints and differentiable simulations, extending to richer geometric formats such as meshes, and enforcing contact-aware reasoning. We also plan to expand MechBench with additional mechanisms (e.g., linkages, cam–follower systems) to support broader and more generalizable motion prediction.

\bibliographystyle{IEEEtran}
\bibliography{ICRA2026_conference}

\end{document}